# Offensive Hebrew Corpus and Detection using BERT

Nagham Hamad
*Birzeit University*
Palestine
nhamad@birzeit.edu

Mustafa Jarrar
*Birzeit University*
Palestine
mjarrar@birzeit.edu

Mohammad Khalilia
*Birzeit University*
Palestine
mkhalilia@birzeit.edu

Nadim Nashif
*7amleh Center*
Palestine
nadim@7amleh.org

*Abstract*—Offensive language detection has been well studied in many languages, but it is lagging behind in low-resource languages, such as Hebrew. In this paper, we present a new offensive language corpus in Hebrew. A total of 15,881 tweets were retrieved from Twitter. Each was labeled with one or more of five classes (abusive, hate, violence, pornographic, or none offensive) by Arabic-Hebrew bilingual speakers. The annotation process was challenging as each annotator is expected to be familiar with the Israeli culture, politics, and practices to understand the context of each tweet. We fine-tuned two Hebrew BERT models, HeBERT and AlephBERT, using our proposed dataset and another published dataset. We observed that our data boosts HeBERT performance by 2% when combined with $D_{OLaH}$. Fine-tuning AlephBERT on our data and testing on $D_{OLaH}$ yields 69% accuracy, while fine-tuning on $D_{OLaH}$ and testing on our data yields 57% accuracy, which may be an indication to the generalizability our data offers. Our dataset and fine-tuned models are available on GitHub and Huggingface.

*Index Terms*—Offensive, Deep Learning, Hate speech, Hebrew, Pre-trained model.

## I. Introduction

The amount of content published on social media is massive and cannot be moderated manually [1]. This has led to widespread of offensive language, adding pressure on social media platforms to moderate and monitor the content posted by the users [2] [3]. Governments, human rights organizations, and social network platforms can benefit from the automatic detection of offensive and hate language.

The advancements in Natural Language Processing (NLP) have opened the door for new technologies that can automate offensive language detection [4]. Many studies focused on classifying texts into offensive [3], abusive [5], irony [4], cyberbullying [6], Sentiment Analysis [7] and opinion mining [8]. Others also suggest to use natural language understanding techniques (e.g., named-entity recognition [9] and word-sense disambiguation [10]) for detecting cybercrimes from social media platforms [11].

These studies focused on building and annotating corpora collected from social media platforms such as Twitter, Facebook, and YouTube. However, determining what is offensive in a given language is a challenging task since it highly relies on one's knowledge and familiarity with the linguistic and cultural aspects of that language [12] [13].

The problem is even more challenging when dealing with the colloquial text given the wide variety of Arabic dialects [14] [15] [16].

The offensive content in Hebrew is wide-spreading on social media, especially against Arabs and Palestinians (see Figure 1). Limited attention is given to this content as Hebrew lacks resources for offensive language detection. To the best of our knowledge, there are only two small datasets in Hebrew for offensive language detection. The first dataset [17] consists of 1,489 posts and comments collected from Facebook, but it is not publicly available. The second dataset, $D_{OLaH}$, consists of 2,000 Facebook posts [18]. A combination of both datasets, with a small extension, was released recently [19].

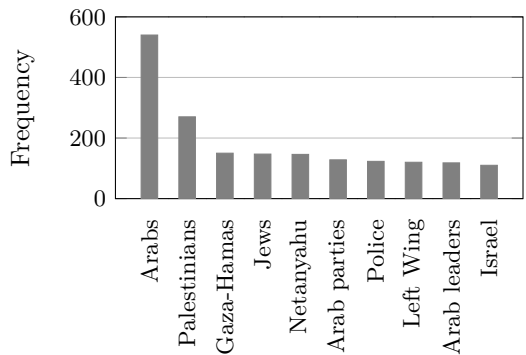

Figure 1. The top 10 targets in the dataset.

This paper presents a new offensive language in Hebrew that consists of 15,881 tweets. The dataset is used to fine-tune HeBERT [20] and AlephBERT [21] models to detect offensive language. We performed multiple experiments that combined our dataset with $D_{OLaH}$. We observed that our data boosts HeBERT performance by 2% when combined with $D_{OLaH}$. Fine-tuning AlephBERT on our data and testing on $D_{OLaH}$ yields 69% accuracy, while fine-tuning on $D_{OLaH}$ and testing on our data yields 57% accuracy, which may be an indication to the generalizability our data offers. Our dataset, fine-tuned models, and the source code are available publicly on GitHub[1], HuggingFace[2] and SinaResources[3].

---

[1] https://github.com/SinaLab/OffensiveHebrew
[2] https://huggingface.co/SinaLab/OffensiveHebrew
[3] https://sina.birzeit.edu/resources/

The main contributions of this paper are:
1) A Hebrew dataset of 15,881 tweets for detecting offensive language, manually annotated with a label, targets, topics, and offensive phrases.
2) A fine-tuned HeBERT and AlephBERT models using different data configurations.

The rest of the paper is organized as follows: Section II reviews the related studies and datasets. Section III explains the methodology we used to collect and annotate our dataset. Section IV illustrates our model architecture. Section V describes the experiments and summarizes the results. Section VI concludes the paper. Finally, Section VII discusses the limitations and future directions of our work.

## II. Related Work

This section reviews work related to offensive language detection. Since the research on Hebrew offensive language detection and datasets is limited, we review work related to Arabic as an other Semitic language [22].

Shared tasks have accelerated the progress on some of the challenging problems. For instance, as part of the SemEval-2019 Task 6, [23] presented a shared task to identify and categorize offensive language in social media. The task was divided into three sub-tasks: 1) Sub-task A for offensive language identification, 2) Sub-task B for automatic categorization of offense types, and 3) Sub-task C for offense target identification. The shared task was based on the Offensive Language Identification Dataset (OLID) [24], which contains about 14K tweets in English. The data was collected from Twitter based on a list of pre-defined keywords. A three-level hierarchical annotation process was used: 1) label a tweet as offensive or not, 2) determine the type of offense if the tweet contains insults or threats, and 3) identify the targeted individual(s) or group(s).

Classical machine learning and deep learning models were evaluated on this task. Ensembles-based approaches achieved the best results. For example, [24] compared different models that combined uni-gram features and Support Vector Machines (SVM) with Convolution Neural Networks (CNN) and Bidirectional Long-Short Term Memory (BiLSTM) on the three sub-tasks. CNN outperformed all models on sub-tasks A and B with F1-score 80% and 69%, respectively. However, the three models (SVM, CNN, and BiLSTM) achieved similar results on sub-task C with 47% F1-score.

SemEval-2020 Task 12 offered the same three sub-tasks as proposed in [24], with the exception of sub-task A which was extended to include five languages (Arabic, Danish, English, Greek, and Turkish). [25] took advantage of transfer learning between related languages by fine-tuning multilingual Bidirectional Encoder Representations from Transformers (BERT) on the Greek, Danish, and Turkish datasets achieving a macro F1-score of 85%, 79%, and 80% for each language, respectively.

In the last few years, Arabic hate speech detection started to gain more attention. This is because of the increased number of people using social media [26]. Most of the published work focuses on building and annotating Arabic hate speech corpora. [12] proposed to use a list of obscene terms to search for offensive tweets. The list was extracted from tweets that were collected during March 2014 using Twitter streaming API. Moreover, some patterns are used to search based on vocative cases such as "you son(s) of" to collect words appearing after these patterns. The final list included 288 offensive words and phrases and 127 hashtags that are employed in an online tweet aggregator to filter out obscene pages in an online tweet aggregator TweetMogaz [27]. This list was used to generate a dataset containing 1,100 manually labeled tweets. In addition, they extracted a list of users, also known as tweeps, from the collected tweets by determining people who frequently use obscenity terms. In addition, [12] released Aljazeera Deleted Comments dataset containing 32K user comments collected from a well-known Arabic news website "aljazeera.net". According to the Aljazeera community rules, a user comment is rejected if it is a personal attack, racist, sexist, or offensive.

[28] introduced the first religion-related Arabic dataset for hate speech. The dataset contains data for six religions (Judaism, Atheism, Shia, Christianity, Sunni, and Islam). Using the Twitter's search API, 6,000 tweets (1,000 for each religion) were collected in November 2017. The tweets were extracted using a list of keywords defined by the authors without using any religious slurs to avoid bias in the data. The data was used for training a binary lexicon-based classifier, SVM using n-gram features, and Gated Recurrent Units (GRU). The models were evaluated on 600 tweets (100 for each religion) collected from January 2018 to detect hate or none hate tweets. The GRU model outperformed all other models with 79% accuracy and 77% F1-score.

Other researchers used hybrid models to achieve better accuracy. For example, [29] extracted a dataset using Twitter API based on a list of hashtags that indicate hostile content. As a result, they generated a dataset of 11K tweets classified into six classes: religious, racial, sexism, general hate, and none. Four deep learning models (LSTM, CNN+LTSM, GRU, and CNN+GRU) were used with the SVM model as a baseline. The experiments showed that a model combining CNN with LSTM and CNN with GRU achieves the best performance with 73% F1 score.

[30] published a new Arabic Hate Speech dataset (ArHS) collected based on a lexicon using Twitter4J API. They labeled 9,833 tweets that are classified into five classes: misogyny, racism, religious discrimination, abusive, and normal. They found that a CNN-LSTM model performs the best with 73% accuracy for binary classification, 67% for ternary classification, and 65% for multi-class classification.

[31] used the same dataset provided in SemEval-2020 [32] Arabic offensive language task. This dataset contains 10K tweets labeled for detecting hate speech and offensive language. There were two tasks to classify tweets into: 1) offensive or non-offensive, and 2) hate or not hate. Based on the definition of the task and the annotated data, they assumed that if a sentence contains hate speech, then it is offensive. Utilizing this correlation, they evaluated multiple multi-task learning models including BiLSTM, CNN-BiLSTM, and BERT. They showed that CNN combined with BiLSTM outperformed the other models with 90% F1-score for offensive language detection and 73% for hate speech detection.

Additionally, [3] proposed an automated emoji-based approach that depends on emojis to extract a large number of offensive tweets. They collected emojis from [33] [34] and others from "emojipedia.org". Based on the collected emojis, tweets are extracted between June 2016 and November 2017. After removing duplicates and short tweets, the final size of their dataset was 12,698 tweets. The tweets were annotated manually into abusive, offensive, hate, vulgar, and violence. Different machine learning methods were used such as SVM with character n-gram and word n-gram, AraBERT [35], multilingual BERT [36], XLM-RoBERTa [37] and QARiB [38]. The results show that monolingual models such as AraBERT outperformed the multilingual models for detecting offensive and hate speech language with 92% accuracy and 80% F1-score.

For Hebrew, limited research was conducted. [17] used Facebook Graph API to collect posts and comments of 130 Members of Knesset (i.e., parliament) between 2014 and 2016. They collected 5.37M comments, but only 1,489 comments were manually annotated, resulting in an imbalanced dataset (1216 non-abusive, 266 abusive, and 7 unknown comments). To address the data imbalance issue, the authors compiled a list of 683 abusive terms, based on which they retrieved additional comments containing at least one of the abusive word. They added new 950 comments to a total of 1,216 abusive comments. They trained an SVM classifier using word and character n-grams features, which resulted in 83% accuracy. Nevertheless, this dataset is not publicly available. A more recent dataset for Hebrew was developed by [18], who retrieved 2,026 comments through Facebook Graph API based on a set of keywords. The comments were then labeled as offensive and non-offensive. The final dataset contains 1,205 offensive and 821 non-offensive comments, which were then divided into 80% for training and 20% for testing. They trained different machine learning models including Random Forest (RF), SVM, Logistic Regression, and XGBoost using bag-of-words features, in addition, to fine-tuning HeBERT [20]. The two best performing models were HeBERT and random forest with 77.5% and 78.3% accuracy, respectively. More recently, [19] combined the two datasets from ( [17], [18]), and augmented the data with additional offensive language, resulting in a dataset with 5,217 comments. They evaluated the combined dataset using mBERT model and achieved 83.3% accuracy

This paper presents a new open-source Hebrew offensive language dataset comprising 15,881 tweets to encourage more research toward offensive language detection in Hebrew.

### III. THE DATASET

This section presents our progress in developing the dataset and the annotation guidelines.

#### A. Data Collection

The data was collected between December 2020 and January 2021 using Twitter API. We used a list of Hebrew keywords to retrieve a list of candidate tweets. We started with a list of 147 keywords that we assumed to be likely used in offensive tweets. The list was reduced to 55 keywords by removing inflections. Using these 55 keywords, we collected 15,881 tweets free of duplicates and retweets. The list of keywords is available on our GitHub repository[4], and Table I shows a sample of these keywords with English translation.

Table I
SAMPLE KEYWORDS USED TO QUERY TWEETER

| Term in Hebrew | English Translation |
|---|---|
| איסלאמיסט | Islamist |
| מחבל | Terrorist |
| מתנחל | Settler |
| נכבה | Nakba |
| עזה | Gaza |
| אינתיפאדה | Intifada |
| הריסה | Demolition |
| רקטות | Rockets |
| אנטישמי | Anti-semitic |
| התלהבות | Hamas |
| ימות | Will Die |
| לֶאֱנוֹס | Rape |
| תן לזה להישרף | Let it burn |
| הֶרֶג | Killing |
| ערבי | Arabic |
| מוסלמי | Muslim |
| לִשְׁחוֹט | Slaughter |
| לשרוף | Burn |
| מוחמד | Muhammad |

Table II
TWEET STATISTICS IN NUMBER OF CHARACTERS

| Stat | Value |
|---|---|
| Min length | 3 |
| Max length | 100 |
| Average | 25.3 |
| Std. Dev. | 14.9 |

Table II presents basic statistics of the dataset. Most of the offensive tweets found in the data are targeting Arabs and Palestinians (See Figure 1).

---
[4]https://github.com/SinaLab/OffensiveHebrew

## B. Dataset Annotation and Guidelines

Three graduate students were carefully selected and trained by an expert. All of them are Palestinians living in Haifa and are Arabic-Hebrew bilingual speakers. In addition, the students were selected based on their familiarity with Israeli politics and culture. Each student annotated about 5,300 tweets then all annotations were reviewed by the expert. For each tweet, the annotators used the following guidelines:

- **Class**: each tweet was labeled with one or more of the offensive labels (hate, abusive, violence, pornographic, or none). The definitions of these offensive classes are presented in Table III.
- **Target**: for each offensive tweet, the offended *targets* are extracted, which are the people or group(s) that the tweet is offending. If the target(s) is not mentioned explicitly in the tweet, then the annotators were asked to infer it. In ambiguous cases, the tweets are considered *UNT* (un-targeted).
- **Topic**: for each offensive tweet, the topic(s) of offense are identified (war, elections, occupation, politics, rape, etc.).
- **Phrase**: for each offensive tweet, the *offensive phrase(s)* mentioned in the tweet were extracted. NULL is used if no specific offensive phrase is found in the tweet.

Table V provides examples of offensive tweets and their annotations.

Table III presents the counts of each sub-class of offense. There are 1200 tweets labeled with one (or more) sub-classes of offense in the dataset, while 14681 are not offensive. Since our dataset is imbalanced (1200:14681), we plan to enrich it with more offensive.

## C. Dataset Review

We could not conduct a full inter-annotator agreement evaluation because the students who participated in the annotation were not available. Nevertheless, we recruited an external Arabic-Hebrew bilingual speaker familiar with the Israeli culture to review part of the annotations. We selected 3355 tweets (21% of the dataset), as the following: all of the 1314 tweets that are annotated with one of the offensive classes (Abusive, Hate, Violence, Pornographic), in addition to randomly selecting 2041 tweets that are NOT offensive. These tweets were then given to the reviewer to review and revise.

Due to the subjective and complex nature of deciding what is offensive (see section VII), we cannot consider the revised annotations to be more accurate. Therefore, we compare between the original and revised annotation by counting the number of changes to each label. Table IV presents the number of changes made by the reviewer to each label. For example, among the 631 Hate tweets in the dataset, 26 changes were made by the reviewers. The total number of changes made on all offensive sub-classes is 73, which indicates a high agreement between the annotators and the reviewer. The number of changes made to the Target is the biggest (88 changes to 1200 tweets). We believe this is because the values of these labels are open textual values.

## IV. Model Architecture

We used the proposed dataset to fine-tune a transformer-based model, as shown in Figure 2. Our model encodes the Hebrew text and generates representations using a pre-trained Hebrew BERT model. We experimented with two pre-trained BERT models, HeBERT and AlephBERT. HeBERT [20] was trained on two datasets: (*i*) a Hebrew version of the Open Super-large Crawled ALMAnaCH coRpus (OSCAR), which contains about ∼9.8 GB of data, including 1 billion words and over 20.8 million sentences [39], and (*ii*) a Hebrew dump of Wikipedia (∼650 MB of data, including over 63 million words and 3.8 million sentences). AlephBERT [21], was trained on three datasets: (*i*) OSCAR corpus, (*ii*) Twitter (∼6.9 GB of data, including 774 million words and over 71.5 million sentences), and (*iii*) all Hebrew texts extracted from Wikipedia (∼1.1 GB of data, including 127 million words and over 6.3 million sentences). Next, we fed the output from BERT into a dense linear layer which performed the binary classification.

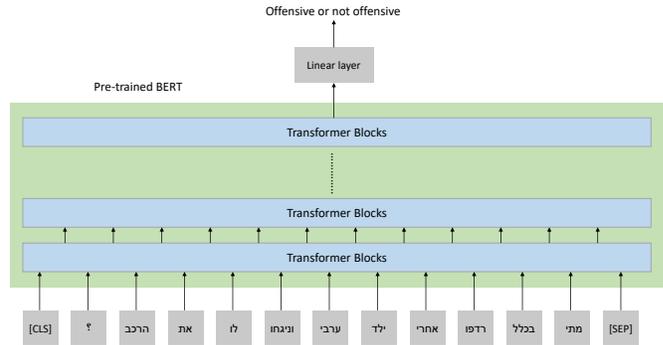

Figure 2. Model architecture.

## V. Experiments and Results

### A. Dataset Preparation

This section illustrates the preparation of the datasets used for training. First, for offensive detection purposes, we combined all sub-classes shown in Table III under the parent class to convert the problem into a binary classification task. In other words, we mapped the labels (hate, abusive, violence, and pornographic) into one label we called offensive. This resulted in a highly imbalanced dataset with 14,681 tweets labeled as *none offensive* and 1,200 tweets labeled as *offensive*. To produce a more balanced dataset, we combined the 1,200 *offensive* tweets with a random sample of 1,300 selected from the *none*

Table III
OFFENSIVE SUB-CLASSES DEFINITIONS AND NUMBER OF TWEETS PER SUB-CLASS.

| Class | Sub-Class | Definition | Count |
|---|---|---|---|
| Offensive | Abusive | If the tweet contains direct or implicit insults using vulgar or street words. | 124 |
| | Hate | If the tweet contains criticism, attack, or degrade, directly or implicitly, because of race, color, religion, nationality, or gender. | 631 |
| | Pornographic | If the tweet promotes or invites any pornographic or sexual arousal. | 4 |
| | Violence | If the tweet endorses an act that involves physical harm towards any party, regardless of the reason. | 454 |
| Not offensive | Not | If the tweet does not contain any offensive language. | 14,681 |
| | | Total | 15,881 |

Table IV
REVIEW: NUMBER OF CHANGES MADE DURING THE REVIEW PHASE

| Label | Total Number | changes |
|---|---|---|
| Hate | 631 | 26 |
| Abusive | 124 | 21 |
| violence | 454 | 26 |
| Porographic | 4 | 0 |
| **Total (offensive)** | | **73** |
| Target | 1202 | 88 |
| Phrase | 1202 | 21 |
| Topic | 1202 | 21 |

*offensive* class, resulting in a more balanced dataset of 2,500 tweets. We split the 2,500 tweets into training (70%), validation (10%), and test (20%) sets. Although we conducted several experiments on our dataset, we also conducted other experiments by combining our dataset with the dataset published in [18], the Offensive language in Hebrew (OLaH), hereafter we refer to it as $D_{OLaH}$. $D_{OLaH}$ contains 2,026 comments (1,205 *none offensive* comments and 821 *offensive* comments) collected from Facebook. In total, we produced eight dataset configurations described in Table VI. We briefly describe those configurations below:

- The datasets $D_1$, $D_2$, and $D_3$ uses 2,500 records from our dataset, but we varied the number of examples from $D_{OLaH}$. $D_1$ contains only our dataset. $D_2$ contains $D_1$ and 1,013 tweets from $D_{OLaH}$. $D_3$ contains $D_1$ and 2,026 tweets from $D_{OLaH}$.
- The datasets $D_4$, $D_5$, and $D_6$ use the same dataset in $D_{OLaH}$, but we varied the number of examples from our dataset. $D_4$ contains all 1,418 records from $D_{OLaH}$. $D_5$ contains $D_4$ and 1,250 tweets from our dataset. $D_6$ contains $D_4$ and 2,500 tweets from our dataset.
- The datasets $D_1$, $D_2$, $D_3$, and $D_7$ use 500 tweets from our dataset for testing.
- The datasets $D_4$, $D_5$, $D_6$ and $D_8$ use 405 tweets from $D_{OLaH}$ for testing.

*B. Experimental Settings*

We used the eight dataset configurations to fine-tune the HeBERT and AlephBERT models. The maximum number of epochs was set to 10 and batch size, $B = 8$, for all models. We used Adam optimizer with a learning rate $\eta = 1e^{-5}$. The maximum input sequence length was 128.

*C. Results*

In this section, we present the performance of the HeBERT and AlephBERT pre-trained models on the datasets described in Section V-A. We will first examine the results on the $D_4$ dataset, which is the original dataset found in [18]. On $D_4$, HeBERT achieved an accuracy of 79%, which is similar to the results reported in [18]. However, AlpehBERT significantly outperformed HeBERT on this dataset by 7% as shown in Table VI. Overall, AlpehBERT outperformed HeBERT on most datasets. A similar conclusion was reported in [21], where AlpehBERT outperformed HeBERT on sentence-level tasks. Although HeBERT and AlpehBERT share the same architecture, the difference in performance can be attributed to two factors: ($i$) AlephBERT has a vocabulary size of 52K, while HeBERT vocabulary size of 30K, ($ii$) AlephBERT is trained on three datasets, Oscar-Wikipedia-Tweets, while HeBERT is trained on two datasets, Oscar-Wikipedia.

We noticed that our training data boosts the HeBERT model performance when combined with $D_{OLaH}$ dataset. Notice that HeBERT when trained on the $D_4$ data configuration (train, test, and validation from $D_{OLaH}$) achieves 79% accuracy. Adding our data to the same model as in the $D_5$ and $D_6$ configuration boosts the model performance by 2% to 81% accuracy.

The behavior is different when using AlpehBERT. When using AlephBERT, we observed that mixing our dataset with $D_{OLaH}$ does not improve the performance of the model on either test datasets. On the contrary, a model trained on the combined datasets suffers from performance degradation on all test datasets. For instance, $D_2$ and $D_3$ were trained on combined data, but performance on our test data decreased by 5% compared to the model trained on $D_1$. We notice a similar behavior with $D_5$ and $D_6$ where the performance on the $D_{OLaH}$ test data decreased by 7% and 4%, respectively, compared to the results produced by the model trained on $D_4$. This type of performance degradation could be due to multiple reasons. First, the datasets may come from different data distributions, different domains, and different sources. Second and most importantly, it is very likely that the annotation guidelines between the two datasets are different, e.g., what is offensive for one community might not be offensive for another.

TABLE V
EXAMPLE OF ANNOTATED TWEETS PER CLASS.

| Tweet | English Translation | Annotations |
|---|---|---|
| גירוש, הריסת בתים, מאסרי עולם ללא אפשרות חנינה, אחרת יהיה עוד יותר גרוע. להתייחס אליהם כמו אל מחבלים. | Deportation, demolition of houses, life sentences without the possibility of amnesty, otherwise it will be even worse. Treat them like terrorists. | **Class**: Violence, Hate<br>**Target**: Palestinians<br>**Topic**: punish Palestinians<br>**Phrase**: Deportation, Demolition of houses, life sentences, terrorists |
| אין כבר הרתעה.לא מפחדים מהמשטרה.אני חושב שהגיע הזמן על פי מראות ההפגנות וההתפרעויות בימים האחרונים,כמו שאמר פעם רבין בתחילת האינתיפאדה: „לשבור להם את העצמות".פה יש כבר אינתיפאדה של התפרעויות. | There is no more deterrence. We are not afraid of the police. I think the time has come to face the demonstrations and riots, as Rabin once said at the beginning of the intifada: "to break their bones." There is already an intifada of riots here. | **Class**: Violence, Hate<br>**Target**: Palestinians<br>**Topic**: Demonstrations<br>**Phrase**: Break their bones |
| @Onetruth011 ימח שמה וזכרה של אילנה דיין. העיתונאית הכי מנוולת ושקרנית שאני מכיר. ממש מרשעת. | May the name and memory of Ilana Dayan be remembered. The most depraved and lying journalist I know. Really sinister. | **Class**: Abusive<br>**Target**: Ilana Dayan<br>**Topic**: Journalism<br>**Phrase**: Sinister, Depraved, Lying |
| @judash0 פרצופו האמיתי של אבי ביטון נחשף לעיני כל. מדובר בשמאלני, אנטי ציוני, עוכר ישראל, בוגד שממומן ע״י הקרן החדשה להפיל את שלטון הימין ולהעלות את המפלגות הערביות לשלטון כדי להוביל למדינת כל אזרחיה | Avi Beaton's true face clear now. This is a leftist, anti-Zionist, oppressor of Israel, a traitor who is financed to overthrow the right-wing government and bring the Arab parties to power in order to lead to a state for all its citizens. | **Class**: Hate, Abusive<br>**Target**: Avi Bitton, Arab Parties<br>**Topic**: politics<br>**Phrase**: Traitor, Anti-Zionist |
| ה. @rabea_bader ואו לא רלוונטי אם אתה דרוזי, סורי, אנטי ציוני ומגעיל שכמוך. | @rabea_bader is irrelevant if you are Druze, Syrian, anti-Zionist and disgusting like you. | **Class**: Hate, Abusive<br>**Target**: Rabea Bader, Druze, Syrian<br>**Topic**: Racism<br>**Phrase**: disgusting, anti-Zionist |
| @Ahmad_tibi אתה לפחות לא משקר - היית ונשארת לאומן ערבי שרוצה בחורבן ישראל כמדינה יהודית. | @Ahmad_tibi At least you're not lying - you were and remain an Arab nationalist who wants the destruction of Israel as a Jewish state. | **Class**: Hate<br>**Target**: Ahmad Tibi<br>**Topic**: Political views<br>**Phrase**: |

The final observation is related to the model's generalizability. Although the best performing model on our data achieved only 68% compared to the 86% accuracy on the $D_{OLaH}$ dataset, we noticed that our data generates a more generalizable model. That is what the data configurations $D_7$ and $D_8$ examine. The $D_7$ data configuration is trained on $D_{OLaH}$ dataset and evaluated on our test dataset, resulting in 57% accuracy using AlephBERT. On the other hand, the $D_8$ configuration uses our training data and is evaluated on $D_{OLaH}$ test data, resulting in 69% accuracy using AlpehBERT. This may indicate that our data generalizes better on new unseen datasets like $D_{OLaH}$.

VI. CONCLUSION

In this paper, we presented our work on developing a new Hebrew offensive language dataset. We collected the data from Twitter based on specific keywords, then annotated the corpus manually using four types of tags.

We fine-tuned two Hebrew BERT models, HeBERT and AlpehBERT, and overall we found that AlpehBERT outperformed HeBERT on six out of the eight data configurations. We also concluded that the model fine-tuned on our data is more generalizable than the model fine-tuned on $D_{OLaH}$. We found that the model fine-tuned on our data achieved 69% accuracy on $D_{OLaH}$ test dataset, while a model fine-tuned on $D_{OLaH}$ achieved 57% accuracy on our test dataset. Although combining our data with $D_{OLaH}$ did not improve the performance of AlpehBERT, we observed our data added a boost to HeBERT performance by 2% when combined with $D_{OLaH}$. Additionally, the model fine-tuned on $D_{OLaH}$ using AlpehBERT outperformed the HeBERT results reported in [18].

Table VI
Eight dataset combinations, and two test sets.

| Data set | # of training examples | | Validation data | Test data | Accuracy | |
|---|---|---|---|---|---|---|
| | Our data | $D_{OLaH}$ | | | He BERT | Aleph BERT |
| $D_1$ | 1,750 | 0 | 250 (ours) | 500 (ours) | 63% | 68% |
| $D_2$ | 1,750 | 1,013 | 250 (ours) | 500 (ours) | 58% | 63% |
| $D_3$ | 1,750 | 2,026 | 250 (ours) | 500 (ours) | 61% | 63% |
| $D_4$ | 0 | 1,418 | 203 ($D_{OLaH}$) | 405 ($D_{OLaH}$) | 79% | **86%** |
| $D_5$ | 1,250 | 1,418 | 203 ($D_{OLaH}$) | 405 ($D_{OLaH}$) | **81%** | 79% |
| $D_6$ | 2,500 | 1,418 | 203 ($D_{OLaH}$) | 405 ($D_{OLaH}$) | **81%** | 82% |
| $D_7$ | 0 | 2,026 | 203 ($D_{OLaH}$) | 500 (ours) | 60% | 57% |
| $D_8$ | 2,500 | 0 | 250 (ours) | 405 ($D_{OLaH}$) | 64% | 69% |

## VII. Limitations and Future Work

Most of the published work, including ours, collected data primarily from Twitter and others from Facebook. We plan to collect more data from different sources and cover periods spanning key events in the ongoing conflicts and hate content in Hebrew. This may help us maximize the size of offensive language and thus produce a more balanced dataset.

As stated earlier, offensive language annotation is challenging. In fact, to aim for high-quality annotation, annotators should ideally be Hebrew native speakers and are familiar with the Palestinian and Israeli cultures and politics. Finding annotators with these criteria has been challenging and labeling high-quality data requires more than one annotator. We plan to revisit our annotations and conduct an evaluation to measure the inter-annotator agreement.